\documentclass[10pt,twocolumn,letterpaper]{article}

\usepackage{iccv}
\usepackage{times}
\usepackage{epsfig}
\usepackage{graphicx}
\usepackage{amsmath}
\usepackage{amssymb}

\usepackage{helvet} 
\usepackage{courier}  
\usepackage{url}  
\usepackage{color}
\usepackage{amsfonts}
\usepackage{bbm}
\usepackage{float}
\usepackage[ruled,lined]{algorithm2e}
\usepackage{epsfig}
\usepackage{booktabs}
\usepackage{color, colortbl}

\usepackage[colorlinks=true,breaklinks=true,bookmarks=false]{hyperref}

\iccvfinalcopy 

\def\iccvPaperID{1417} 

\ificcvfinal\fi

\begin{document}

\title{Single-Stage Multi-Person Pose Machines}

\author{\normalsize{Xuecheng~Nie$^1$} \quad  \qquad \normalsize{Jianfeng~Zhang$^1$} \quad  \qquad \normalsize{Shuicheng~Yan$^{1,2}$}  \quad \qquad \normalsize{Jiashi~Feng$^1$}\\
	\small{$^{1}$Department of Electrical and Computer Engineering, National University of Singapore, Singapore} \\
	\small{$^{2}$Yitu Technology} \\
	{\small \tt niexuecheng@u.nus.edu}   \ \ {\small\tt elezji@nus.edu.sg} \ \ {\small\tt shuicheng.yan@yitu-inc.cn} \ \  {\small \tt elefjia@nus.edu.sg} 
}

\maketitle
\ificcvfinal\thispagestyle{empty}\fi

\begin{abstract}
   Multi-person pose estimation is a challenging problem. 
   Existing methods are mostly two-stage based\textemdash one stage for proposal generation and the other for allocating poses  to corresponding persons. However, such two-stage methods generally suffer  low efficiency.
   In this work, we present the first \emph{single-stage} model, Single-stage multi-person Pose Machine (SPM),  to simplify the pipeline and lift the efficiency for multi-person pose estimation. To achieve this, we  propose a novel Structured Pose Representation (SPR) that unifies  person instance and body joint position representations. Based on SPR, we develop the  SPM model that  can directly  predict structured  poses for 
   multiple persons 
   in a single stage,  and thus offer a more compact pipeline and attractive efficiency advantage over two-stage methods. In particular, SPR introduces the root joints to indicate different person instances and  human body joint positions are encoded into their  displacements w.r.t.~the roots. To better predict  long-range displacements for some joints, SPR is  further extended to hierarchical representations. Based on SPR, SPM can efficiently perform multi-person poses estimation by simultaneously predicting root joints (location of instances) and body joint displacements via CNNs. Moreover, to demonstrate the generality of SPM, we also apply it to multi-person 3D pose estimation. 
   Comprehensive experiments on benchmarks MPII, extended PASCAL-Person-Part, MSCOCO and CMU Panoptic clearly demonstrate the state-of-the-art efficiency of SPM for multi-person 2D/3D pose estimation, together with outstanding accuracy.
\end{abstract}

\section{Introduction}

\begin{figure}[t!]
	\begin{center}
		\includegraphics[scale=0.52]{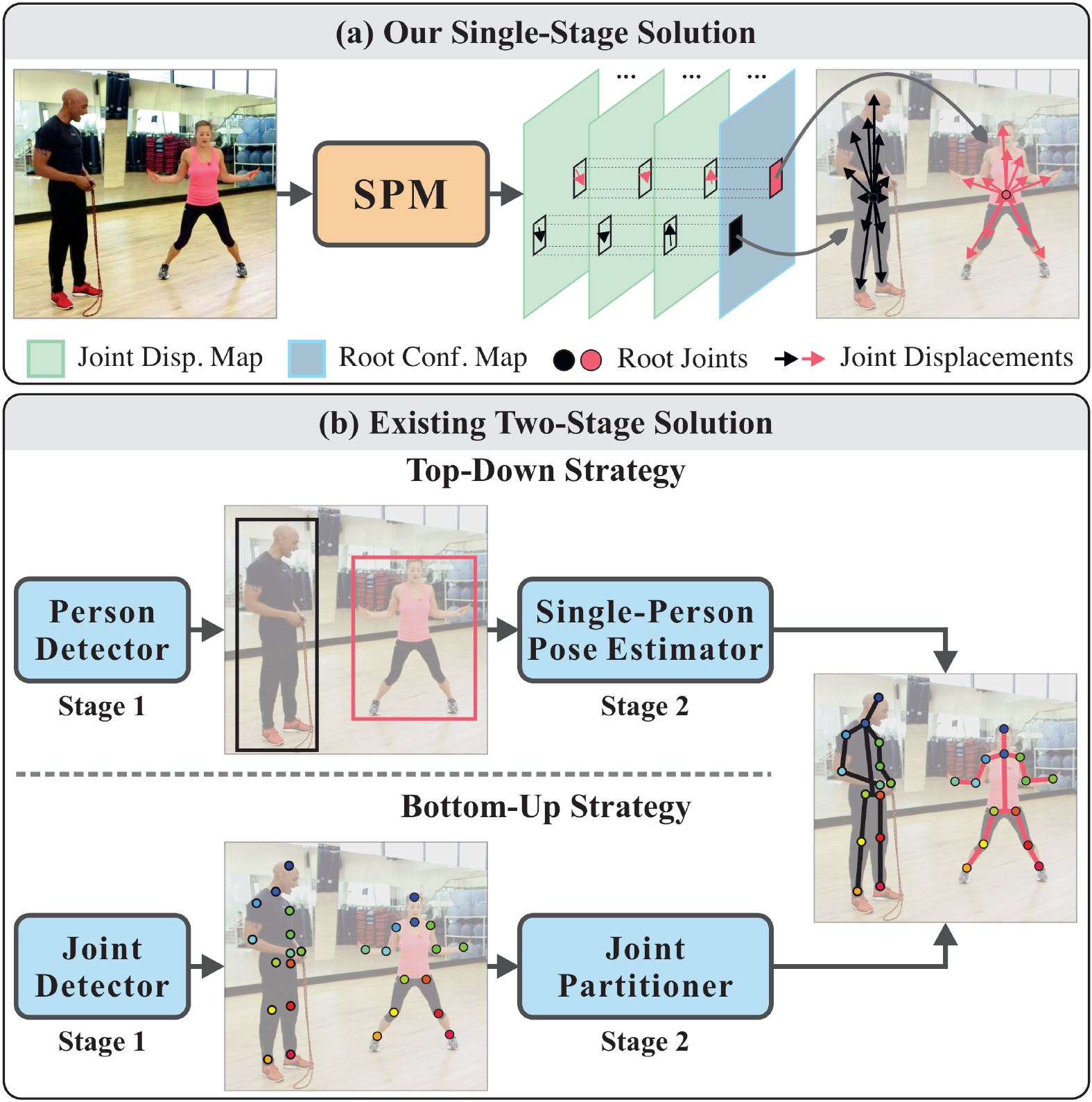}
		\caption{Comparison between (a) our single-stage solution and (b) existing two-stage solution to multi-person pose estimation. The proposed SPM model directly predicts structured poses of multiple persons in a single stage, offering a more compact pipeline and attractive efficiency advantages over two-stage based top-down or bottom-up strategies. See more details in the main text.}
		\label{fig:comp_2stage_1stage}
	\end{center}
	\vspace{-3mm}
\end{figure}

Multi-person pose estimation from a single monocular RGB image aims to simultaneously isolate and locate body joints of multiple person instances. It is a fundamental yet challenging task  with broad applications in action recognition~\cite{cheron2015p}, person Re-ID~\cite{qian2017pose}, pedestrian tracking~\cite{andriluka2010monocular}, etc. 

\begin{figure}[t!]
	\begin{center}
		\includegraphics[scale=0.4]{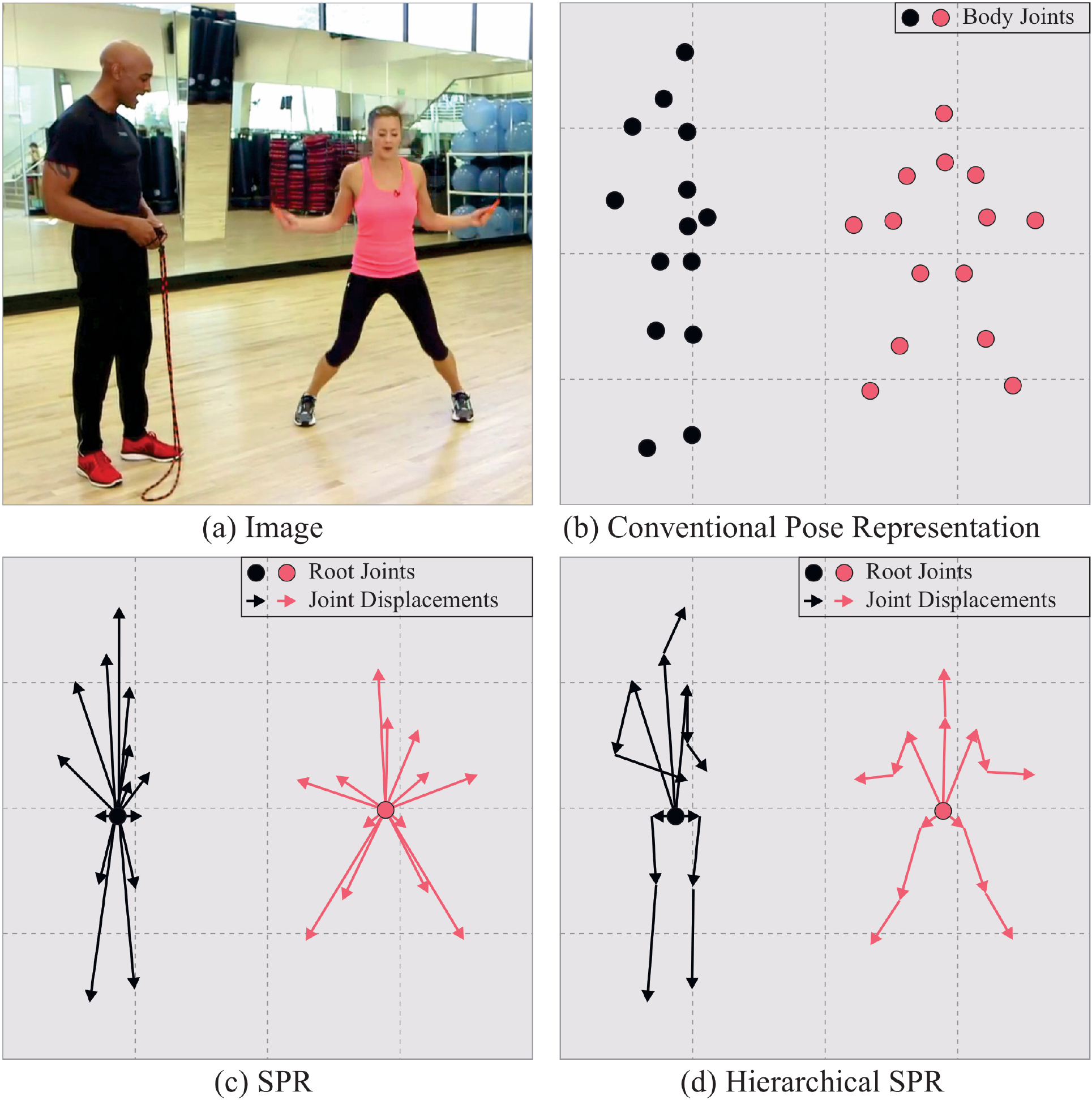}
		\caption{Different pose representations for multiple person instances in image (a). (b) Conventional pose representation with each joint represented by absolute coordinates. (c) Proposed structured pose representation w.r.t.\ root joints. (d) Proposed hierarchical structured pose representation. See more details in  main text.}
		\label{fig:sprs}
	\end{center}
	\vspace{-7mm}
\end{figure}

Existing methods typically adopt two-stage solutions. As shown in Figure~\ref{fig:comp_2stage_1stage} (b), they either follow the top-down strategy~\cite{gkioxari2014using,sun2011articulated,iqbal2016multi,fang16rmpe,dong2019fast,rogez2019lcr} that employs off-the-shelf  detectors to localize person instances at first and then locates their joints individually;  or the bottom-up strategy~\cite{cao2017realtime,hpe:deepercut_eccv16,newell2016associative,hpe:deepcut_cvpr16,mehta2018single} that  locates all the body joints at first and then assigns them to the corresponding  person. Though with high accuracy, these methods are not efficient as they require {two-stage} processing to predict human poses with computational redundancy. We observe that such a requirement mainly comes from the conventional pose representation they adopt. As shown in Figure~\ref{fig:sprs}~(b), absolute positions of allocated body joints separate the position information w.r.t.~person instances and body joints, each of which requires a stage to process, leading to low  efficiency.

To overcome such an intrinsic  limitation, we propose a new Structured Pose Representation (SPR) to unify position information of person instances and body joints. SPR allows to simplify the pipeline for person separation and body joint localization and thus  enables a much more efficient \emph{single-stage} solution to  multi-person pose estimation. In particular, SPR defines a unique identity joint, the \emph{root joint}, for each person instance to indicate its position  in the image. Then, the positions of body joints  are encoded by their displacements w.r.t.\ the root joints. In this way, the  pose of a person instance is represented  together with its location, as shown in Figure~\ref{fig:sprs} (c), making a single-stage solution feasible. To tackle the long-range displacements (\emph{e.g.}  limb joints), we further extend SPR to a hierarchical one by dividing body joints into hierarchies induced from articulated kinematics~\cite{knutzen1998kinematics}. Such a Hierarchical Structured Pose Representation is shown in Figure~\ref{fig:sprs} (d).

Based on SPR, we propose a \emph{Single-stage multi-person Pose Machine} (SPM) model to solve  multi-person pose estimation with compact pipeline and high efficiency. As aforementioned, existing two-stage models isolate different instances and estimate their poses separately. Different from them, SPM maps a given image to multiple human poses represented by SPR in a single-stage manner. As shown in Figure~\ref{fig:comp_2stage_1stage} (a), it simultaneously regresses the root joint positions and body joint displacements, predicting multi-person poses within one stage. We implement SPM with Convolutional Neural Networks (CNNs) based on the state-of-the-art Hourglass architecture~\cite{hpe:hourglass_arxiv15} for learning and inferring root joint position and body joint displacement simultaneously and end-to-end. 

Comprehensive experiments on benchmarks MPII~\cite{andriluka14cvpr}, extended PASCAL-Person-Part~\cite{xia2017joint}, MSCOCO~\cite{lin2014microsoft} and CMU Panoptic~\cite{Joo_2017_TPAMI} evidently demonstrate the high efficiency of the proposed SPM model. In addition, it achieves new state-of-the-art on MPII and extended PASCAL-Person-Part datasets, and competitive performance on MSCOCO dataset. Moreover, it also achieves promising results on CMU Panoptic dataset for multi-person 3D pose estimation. Our contributions is summarized as: 
1) We propose the first \emph{single-stage} solution to multi-person 2D/3D pose estimation. 
2) We propose novel structured pose representations to unify position information of person instances and body joints. 
3) Our model achieves outperforming efficiency with competitive accuracy on multiple benchmarks.

\section{Background}
In this section, we review the state-of-the-art multi-person pose estimation methods based on conventional pose representation. Given an image $I$, multi-person pose estimation targets at estimating human poses $\bar{\mathcal{P}}$ of all the person instances in $I$ via inferring coordinates of their body joints. 
Conventionally, poses are represented as 
\begin{equation}\label{eq:ori_pose_def}
\bar{\mathcal{P}} =  \big\{\mathbf{P}_i^1, \mathbf{P}_i^2, \dots, \mathbf{P}_i^K\big\}_{i=1}^N,
\end{equation}
where $N$ is the number of persons in $I$, $K$ is the number of joint categories, and $\mathbf{P}_i^j$ denotes coordinates of the $j$th body joint from person $i$, where $\mathbf{P}_i^j{=}(x_i^j, y_i^j)$ for 2D case while $\mathbf{P}_i^j{=}(x_i^j, y_i^j,z_i^j)$ for 3D case. To obtain $\bar{\mathcal{P}}$, existing methods typically exploit two-stage solutions, \emph{i.e.} separately predicting positions of person instances 
and their body joints. Based on the processing order,  they  can be divided into two categories: the top-down methods and the bottom-up ones.

A top-down method generates multiple human poses $\bar{\mathcal{P}}$ as follows. It first uses a person detector $f$ to localize and separate person instances, and then conducts  single-person pose estimation using a single-person model $g$ to individually locate body joints for each person instance. Formally,  the process can be summarized as 
\begin{equation}\label{eq:top_down_form}
\begin{split}
f &:  I  \rightarrow \mathcal{B}, \\
g &:  \mathcal{B},I \rightarrow \bar{\mathcal{P}}.
\end{split}
\end{equation}
Here $\mathcal{B}$ denotes person instance localization results that are usually represented by a set of bounding boxes. Following this strategy, for 2D case, Gkioxari \emph{et al.}~\cite{gkioxari2014using} exploited a Generalized Hough Transform framework to detect person instances and then localize body joints via classifying poselets\textemdash the tightly clustered body parts with similar appearances and configurations. 
Iqbal and Gall~\cite{iqbal2016multi} improved the person detector and single-person model via exploiting deep learning based techniques, including Faster-RCNN~\cite{ren2015faster} and convolutional pose machine~\cite{hpe:conv_pose_machine_arxiv16}, to acquire more accurate human poses. Similarly, Fang \emph{et al.}~\cite{fang16rmpe} proposed to incorporate spatial transformer network~\cite{jaderberg2015spatial} and Hourglass network~\cite{hpe:hourglass_arxiv15} to further improve person instance and body joint detections. Papandreou \emph{et al.}~\cite{papandreou2017towards} further improved the top-down strategy via location refinement with predictions of 2D offset vector from a pixel to the corresponding joint. For 3D case, Rogez~\cite{rogez2019lcr} first utilized region proposal network to detect persons of interest and found 3D anchor pose for each detection, then exploit iterative regression for refinement. Dong~\cite{dong2019fast} performed top-down multi-person 2D pose estimation for images from multiple views and reconstructed 3D pose for each person from multi-view 2D poses.

In contrast, to obtain poses $\bar{\mathcal{P}}$, a bottom-up method   first utilizes a body joint estimator $g^\prime$ to localize body joints for all instances, and then estimates the position of each instance and the joint allocation by solving a graph partition problem with the model $f^\prime$, formulated as
\begin{equation}\label{eq:bottom_up_form}
\begin{split}
g^\prime &:  I \rightarrow \mathcal{J}, \mathcal{C} \\
f^\prime &: \mathcal{J}, \mathcal{C} \rightarrow \bar{\mathcal{P}},
\end{split}
\end{equation}
where $\mathcal{J}$ denotes the set of joint candidates and $\mathcal{C}$ the affinities for assigning joint candidates to person instances. In~\cite{hpe:deepercut_eccv16}, Insafutdinov \emph{et al.}~exploited Residual networks~\cite{he2016deep} as the joint detector and defined geometric correlations for allocating body joints, and then performed Integer Linear Programming to partition joint candidates. Cao \emph{et al.}~\cite{cao2017realtime} proposed a real-time model with improved joint correlations via introducing part affinity fields to encode location and orientation of limbs and allocate joint candidates via solving a maximum weight bipartite graph matching problem. Later, Mehta~\cite{mehta2018single} extended~\cite{cao2017realtime} to multi-person 3D pose estimation.  Newell and Deng~\cite{newell2016associative} introduced the associative embedding model followed by a greedy algorithm for allocating body joints. Papandreou \emph{et al.}~\cite{papandreou2018personlab} presented the bottom-up PersonLab model by defining different levels of offsets to calculate association scores and adjust joint positions for grouping joint candidates into person instance and refining pose estimations.

Different from all the previous methods relying on a two-stage pipeline, we present a new  pose representation method that unifies positions of person instances and body joints, enabling a compact and efficient single-stage solution to multi-person 2D/3D pose estimation, as explained  below.

\section{Structured pose representation }
In this section, we elaborate on the proposed Structured Pose Representations (SPR) for multi-person pose estimation. Different from the conventional pose representation in Eqn.~(\ref{eq:ori_pose_def}), SPR aims to unify the position information of person instance and body joint to deliver a single-stage solution for multi-person pose estimation. In particular, SPR introduces an auxiliary joint, the root joint, to denote the person instance position. It is a unique identity joint for a specific person instance. In the following, we illustrate the formulations of SPR in 2D case for simplification, which can be directly extended to 3D case via replacing 2D coordinates with 3D ones. Specifically, we use $(x_i^{\mathrm{r}}, y_i^{\mathrm{r}})$ to denote the root joint position of the $i$th person.  Then the position of the $j$th joint of person $i$ can be defined  as
\begin{equation}\label{eq:pos_factorization}
(x_i^j, y_i^j) = (x_i^{\mathrm{r}}, y_i^{\mathrm{r}}) + (\delta{x}_i^j, \delta{y}_i^j),
\end{equation}
where $(\delta{x}_i^j, \delta{y}_i^j)$ represents the displacement of the $j$th body joint position w.r.t.\ the root joint. Eqn.~(\ref{eq:pos_factorization}) directly establishes the structured relationship between person instance position and body joint position. Thus, we use the Structured Pose Representations to represent human poses with the root joint position and body joint displacements, formulated as
\begin{equation}\label{eq:our_vanilla_pose_def}
\mathcal{P}{=}\big\{(x_i^{\mathrm{r}}, y_i^{\mathrm{r}}), (\delta{x}_i^1, \delta{y}_i^1), (\delta{x}_i^2, \delta{y}_i^2), \dots, (\delta{x}_i^K, \delta{y}_i^K)\big\}_{i=1}^N.
\end{equation}
By the definition in Eqn.~(\ref{eq:our_vanilla_pose_def}), SPR unifies position information of the person instance and the body joint and can be obtained in an efficient single-stage prediction. In addition, SPR can be effortlessly converted back to the conventional pose representation based on Eqn.~(\ref{eq:pos_factorization}). Here, we exploit the person centroid as the root joint of the person instance, due to its stability and robustness in discriminating person instances even with extreme poses. An example of SPR representing multiple human poses is shown in Figure~\ref{fig:sprs} (c).

\vspace{-2mm}
\paragraph{Hierarchical SPR} SPR in Eqn.~(\ref{eq:our_vanilla_pose_def}) may involve long-range displacements between body joints and the root joint due to possible large pose deformation, \eg, wrists and ankles relative to the person centroid, bringing difficulty to displacement estimation by mapping from image representation to the vector domain. Thereby, we propose to factorize long-range displacements into accumulative shorter ones to further improve SPR. Specifically, we divide the root joint and body joints into four hierarchies based on articulated kinematics~\cite{knutzen1998kinematics} by their degrees of freedom and extent of deformation. Here, the  root joint is placed in the first hierarchy; torso joints including neck, shoulders and hips are in the second one; head, elbows and knees are put in the third; wrists and ankles are put in the fourth. Then we can identify joint positions via shorter-range displacements between joints in adjacent hierarchies. For example, the wrist position can be encoded  by its displacement relative to the elbow. Modeling short-range displacements can alleviate the learning difficulty of mapping from image representation to the vector domain and better utilize appearance cues along limbs. 
Formally, for the $j$th joint in the $l$th layer  (\eg, wrist in the 4th layer) and its corresponding $j^\prime$th joint in the $(l{-}1)$th layer (\eg, elbow in the 3rd layer), the relation between their positions $(x_i^{j}, y_i^{j})$ and $(x_i^{j^\prime}, y_i^{j^\prime})$ can be formulated as
\begin{equation}\label{eq:pos_hierar_factorization}
(x_i^{j}, y_i^{j}) = (x_i^{j^\prime}, y_i^{j^\prime}) + (\delta{\tilde{x}}_i^j, \delta{\tilde{y}}_i^j),
\end{equation}
where $(\delta{\tilde{x}}_i^j, \delta{\tilde{y}}_i^j)$ denotes the displacement between joints in adjacent hierarchies. According to the articulated kinematics, we can define an articulated path (a set of ordered joints) 
connecting  the root joint to  any body joint. Then, the body joint can be identified  via the root joint position and accumulation of short-range displacements along the articulated path. Namely,   
\begin{equation}\label{eq:pos_hierar_indication}
(x_i^j, y_i^j) =(x_i^{\mathrm{r}}, y_i^{\mathrm{r}}) + \sum_{h \in \mathcal{H}^j{\backslash}\{\mathrm{r}\}}(\delta{\tilde{x}}_i^h, \delta{\tilde{y}}_i^h),
\end{equation}
where $\mathcal{H}^j=\{\mathrm{r}, a^{(1)}, \dots, a^{(m)}, j\}$ represents the articulated path between the root joint and the $j$th body joint and $a^{(n)}$ denotes the $n$th articulated joint on the path. In this way, we propose the Hierarchical Structured Pose Representations to denote a human pose with the root joint position, the short-range body joint displacements between neighboring hierarchies, and the articulated path set $\mathcal{H}$ as 
\begin{equation}\label{eq:our_hierar_pose_def}
\begin{aligned}
\mathcal{P}{=}\big\{(x_i^{\mathrm{r}}, y_i^{\mathrm{r}}), (\delta{\tilde{x}}_i^1, \delta{\tilde{y}}_i^1), (\delta{\tilde{x}}_i^2, \delta{\tilde{y}}_i^2), \dots, (\delta{\tilde{x}}_i^K, & \delta{\tilde{y}}_i^K) \big\}_{i=1}^N, \\ 
& \text{\quad given } \mathcal{H}.
\end{aligned}
\end{equation}
Similar to SPR, hierarchical SPR defined in Eqn.~(\ref{eq:our_hierar_pose_def}) also unifies representations of person instance position and body joint position, leading to a single-stage solution to multi-person pose estimation as well. Moreover, hierarchical SPR factorizes displacements between the root joint and long-range body joints, benefiting estimation results for the cases with large body joint displacements. Hierarchical SPR can also be easily converted to SPR and conventional pose representation via Eqn.~(\ref{eq:pos_hierar_indication}). Figure~\ref{fig:sprs} (d) gives an example of Hierarchical SPR for multi-person pose representation.

\section{Single-stage multi-person pose machine}
With SPR, we propose to construct a regression model, termed as Single-stage multi-person Pose Machine (SPM), to map an input image $I$ to the poses of multiple persons $\mathcal{P}$:
\begin{equation}\label{eq:reg_model}
\mathrm{SPM}:I \rightarrow \mathcal{P},
\end{equation}
which tackles the multi-person pose estimation problem in a single-stage manner. Different from two-stage solutions in Eqn.~(\ref{eq:top_down_form}) and~(\ref{eq:bottom_up_form}), SPM only needs to learn a single mapping function. Motivated by recent success of Convolutional Neural Networks (CNNs) in computer vision tasks~\cite{he2016deep,lin2016feature,long2015fully}, we implement SPM with a CNN model. Below we will describe regression targets, network architecture, and  training and inference details of SPM in 2D case for simplification. For 3D case\footnote{We set the camera position as the origin of the 3D coordinate system.}, the same scheme can be exploited with 3D coordinates.

\begin{figure}[t!]
	\begin{center}
		\includegraphics[scale=0.4]{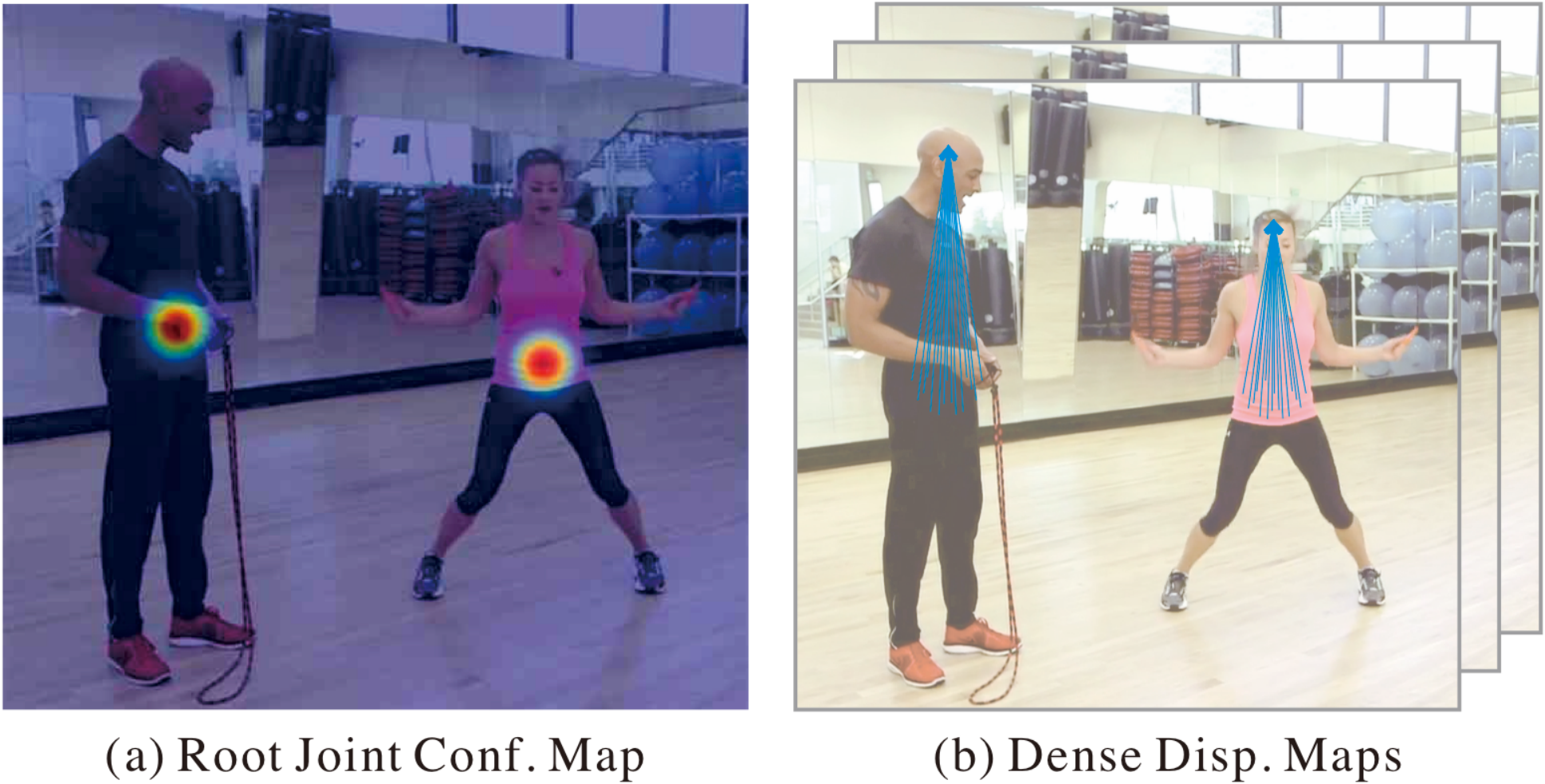}
		\caption{Regression targets of the proposed SPM. (a) Confidence map for root joint. (b) Dense displacement maps for body joints.}
		\label{fig:target_exp}
	\end{center}
	\vspace{-5mm}
\end{figure}

\subsection{Regression targets}\label{sec:reg_target}
Since the root joint $(x_i^{\mathrm{r}},y_i^{\mathrm{r}})$  and body joint displacements $\{(\delta{x}_i^1, \delta{y}_i^1), (\delta{x}_i^2, \delta{y}_i^2),\dots, (\delta{x}_i^K, \delta{y}_i^K)\}$ are respectively in the coordinate and vector domains, we construct different regression targets for the proposed SPM to learn to predict these two kinds of information. 

\vspace{-3mm}
\paragraph{Regression target for root joint position}
According to previous works~\cite{carreira2016human,hpe:deepcut_cvpr16}, it is difficult to directly regress the absolute joint coordinates in an image. To reliably detect root joint positions, we exploit a confidence map to encode probabilities of the root joint of a person instance at each location in the image. The root joint confidence map is constructed by modeling the root joint position as Gaussian peaks. We use $\mathrm{C}^{\mathrm{r}}$ to denote the root joint confidence map and $\mathrm{C}_i^{\mathrm{r}}$ the root joint map of the $i$th person. For a position $(x, y)$ in the given image $I$, $\mathrm{C}_i^{\mathrm{r}}(x, y)$ is calculated by 
\begin{equation*}
\mathrm{C}_i^{\mathrm{r}}(x, y) =  \mathrm{exp}(-\|(x, y) - (x_i^{\mathrm{r}}, y_i^{\mathrm{r}})\|_2^2/\sigma^2),
\end{equation*}
where $(x_i^{\mathrm{r}}, y_i^{\mathrm{r}})$ is the groundtruth root joint position of the $i$th person instance and $\sigma$ is an empirically chosen constant to control the variance of Gaussian distribution, set as $\sigma{=}7$ in our experiments. The root joint confidence map $\mathrm{C}^{\mathrm{r}}$ is an aggregation of peaks of all persons in a single map. Here, we choose to take the maximum of confidence maps rather than their average to maintain distinctions between close-by peaks~\cite{cao2017realtime}, \ie, $\mathrm{C}^{\mathrm{r}}(x, y){=}\max_i\mathrm{C}_i^{\mathrm{r}}(x, y)$. An example of the root joint confidence map is shown in Figure~\ref{fig:target_exp} (a).

\vspace{-3mm}
\paragraph{Regression target for body joint displacement}\label{sec:dis_reg}
We construct a dense displacement map for each  joint. We use $\mathrm{D}^j$ to denote it for joint $j$ and $\mathrm{D}_i^j$ to denote the one for joint $j$ of person $i$. For a location $(x, y)$ in image $I$,  $\mathrm{D}_i^j(x, y)$ is calculated by
\begin{equation*}
\mathrm{D}_i^j(x, y) =  
\begin{cases}
\frac{(\delta{x}, \delta{y})}{Z}   & \text{ if }  (x, y) \in \mathcal{N}_i^{\mathrm{r}}  \\
0                    & \text{ otherwise }
\end{cases}, 
\end{equation*}
\begin{equation*}
(\delta{x}, \delta{y}) = (x_i^j, y_i^j) - (x, y),
\end{equation*}
where $\mathcal{N}_i^{\mathrm{r}}=\big\{(x, y)| \|(x, y)-(x_i^{\mathrm{r}}, y_i^{\mathrm{r}})\|_2^2\leq\tau\big\}$ denotes the neighboring positions of the root joint of person $i$,
$Z{=}\sqrt{H^2 + W^2}$ is the normalization factor, with $H$ and $W$ denoting the height and width of $I$, and $\tau$ is a constant controlling the neighborhood size, set as 7 in our experiments. Then, we define the dense displacement map $\mathrm{D}^j$ for the $j$th joint to be the average for all persons: 
\begin{equation*}
\mathrm{D}^j(x, y) = \frac{1}{M^j}\sum_i\mathrm{D}_i^j(x, y),
\end{equation*}
where $M^j$ is the number of non-zero vectors at position $(x, y)$ across all persons. Figure~\ref{fig:target_exp} (b) shows examples for the constructed dense displacement maps. 
For hierarchical SPR, $\mathrm{D}^j$ is constructed in a similar way, just replacing the root joint with the one in the neighbor hierarchy.

\subsection{Network architecture}
\begin{figure}[t!]
	\begin{center}
		\includegraphics[scale=0.8]{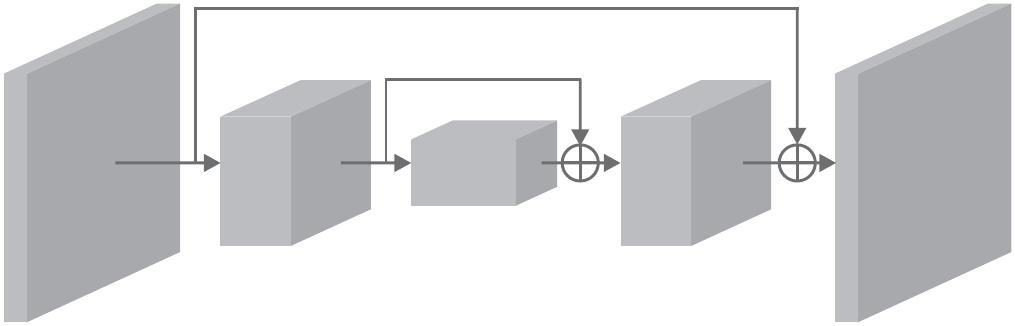}
		\caption{The backbone of SPM: Hourglass network.}
		\label{fig:hg_net}
	\end{center}
\end{figure}

We use the Hourglass network~\cite{newell2016associative}, the state-of-the-art architecture for human pose estimation, as the backbone of SPM. It is a fully convolutional network composed of multiple stacked Hourglass modules. Each Hourglass module, as shown in Figure~\ref{fig:hg_net}, adopts a U-Shape structure that  first decreases feature map resolution to learn abstract semantic representations and then upsamples the feature maps for body joint localization. Additionally, skip connections are added between feature maps with the same resolution for reusing low-level spatial information to refine high-level semantic information. In the original design, the Hourglass network utilizes a single branch to predict body joint confidence maps for single-person pose estimation. In this paper, SPM exploits the confidence regression branch of the Hourglass network to regress confidence maps for the root joint. In addition, SPM extends the Hourglass network via adding a displacement regression branch, to estimate body joint displacement maps. In this way, SPM can produce (Hierarchical) SPR in a single forward pass.

\subsection{Training and inference}\label{sec:train_and_infer}
For training SPM, we adopt  $\ell_2$ loss $\mathcal{L}^{\mathrm{C}}$  and smooth $\ell_1$ loss~\cite{girshick2015fast} $\mathcal{L}^{\mathrm{D}}$ for root joint confidence and dense displacement map regression respectively. Intermediate supervision is applied at all Hourglass modules to avoid gradient vanishing. The total loss $\mathcal{L}$ is the accumulation of weighted sum of $\mathcal{L}^{\mathrm{C}}$ and $\mathcal{L}^{\mathrm{D}}$ across all hourglass modules:
\begin{equation*}
\mathcal{L} = \sum_{t=1}^{T}\big(\mathcal{L}^{\mathrm{C}}(\hat{\mathrm{C}}_{(t)}^{\mathrm{r}}, \mathrm{C}^{\mathrm{r}}) + \beta\mathcal{L}^{\mathrm{D}}(\hat{\mathrm{D}}_{(t)}, \mathrm{D})\big),
\end{equation*}
where $T$ is the number of Hourglass modules, set as $T{=}8$, $\hat{\mathrm{C}}_{(t)}^{\mathrm{r}}$ and $\hat{\mathrm{D}}_{(t)}$ denote the predicted root joint confidence map and dense displacement maps at the $t$th stage, and $\beta$ is a constant weight factor to balance two kinds of losses, set as $\beta{=}0.01$ in our experiments. The overall framework of SPM is end-to-end trainable via gradient backpropagation.

The overall inference procedure for SPM to predict SPR is illustrated in Figure~\ref{fig:comp_2stage_1stage} (a). Given an image, SPM first  produces root joint confidence map $\hat{\mathrm{C}}^{\mathrm{r}}$ and displacement maps $\hat{\mathrm{D}}$ via a CNN. Then, it performs NMS on $\hat{\mathrm{C}}^{\mathrm{r}}$ to generate root joint positions $\big\{(\hat{x}_i^{\mathrm{r}}, \hat{y}_i^{\mathrm{r}})\big\}_{i=1}^{\hat{N}}$, with $\hat{N}$ denoting the estimated number of persons. After that, SPM gets the displacement of the body joint $j$ of person $i$ by $Z{\cdot}\mathrm{D}^{j}(\hat{x}_i^{\mathrm{r}}, \hat{y}_i^{\mathrm{r}})$. Finally, SPM outputs human poses represented by SPRs via combining root joint positions and body joint displacements. For predicting hierarchical SPRs, SPM follows the above procedure to sequentially get joint displacements according to the joint hierarchies  in Eqn.~\eqref{eq:pos_hierar_indication}.  

\section{Experiments}

\subsection{Experiment setup}\label{sec:exp_setup}

\paragraph{Datasets} We evaluate the proposed SPM model for multi-person pose estimation on three widely adopted 2D benchmarks: MPII~\cite{andriluka14cvpr} dataset, extended PASCAL-Person-Part~\cite{xia2017joint} dataset and MSCOCO~\cite{lin2014microsoft} dataset, and one 3D benchmark CMU Panoptic dataset~\cite{Joo_2017_TPAMI}.

MPII dataset contains 5,602 groups of images of multiple persons, which are split into 3,844 for training and 1,758 for testing. It also provides over 28,000 annotated single-person pose samples. Each person is annotated with 16 body joints. We use the official mean Average Precision (mAP) for evaluation on this dataset. The extended PASCAL-Person-Part dataset consists of 1,716 training and 1,817 testing images collected from the original PASCAL-Person-Part dataset~\cite{chen2014detect}, and provides 14 body joint annotations for each person. Similar to MPII, this dataset also adopts mAP as the evaluation metric. MSCOCO dataset contains about 60,000 training images with 17 annotated body joints per person. Evaluations are conducted on the test-dev subset, including roughly 20,000 images, with the official Average Precision (AP) as metric.

CMU Panoptic is a large scale dataset providing 3D pose annotations for multiple people engaging social activities. 
It totally includes 65 videos with multi-view annotations, but only 17 of them are in multi-person scenario and given the camera parameters. We use the front-view captures of these 17 videos in our experiments, which contains 75,552 images in total and are randomly split into 65,552 for training and 10,000 for testing. We following conventions~\cite{mehta2018single,rogez2019lcr} to utilize 3D-PCK@150mm as metric.

\paragraph{Data augmentation} We follow the conventional data augmentation strategies for multi-person pose estimation via cropping original images centered at person centroid to $384{\times}384$ input samples to SPM. For MPII and extended PASCAL-Person-Part datasets, we augment training samples with rotation degrees in $[-40^{\circ}, 40^{\circ}]$, scaling factors in $[0.7, 1.3]$, translation offset in $[-40\mathrm{px}, 40\mathrm{px}]$ and horizontally flipping. For MSCOCO dataset, scaling factors are sampled in $[0.5, 1.5]$ and other augmentation parameters are set the same as MPII and extended PASCAL-Person-Part datasets. For CMU Panoptic dataset, we conduct data augmentation with scale factors in $[0.9, 1.5]$ and set the other augmentation parameters the same as 2D case.

\vspace{-1mm}
\paragraph{Implementation} For MPII dataset, we randomly select 350 groups of multi-person training samples as the validation dataset and use the remaining training samples and all single-person pose images to learn SPM. For MSCOCO dataset, we use the standard training split for training the model. Following conventions~\cite{cao2017realtime,hpe:conv_pose_machine_arxiv16} for multi-person pose estimation, we normalize the input image to CNN with mean 0.5 and standard deviation 1.0 for RGB channels. We implement SPM with Pytorch~\cite{paszke2017pytorch} and utilize RMSprop~\cite{rmsprop2012} as the optimizer with an initial learning rate of 0.003. For MPII dataset, we train SPM for 250 epochs and decrease learning rate by a factor of 2  at the 150th, 170th, 200th, 230th epoch. For extended PASCAL-Person-Part dataset, we fine-tune the model pre-trained on MPII  for 30 epochs. For MSCOCO dataset, SPM is trained for 100 epochs and learning rate is decreased at the 30th, 60th, and 80th epoch by a factor of 2. For CMU Panoptic dataset, we adopt the same training strategy as MPII. Testing is performed on six-scale image pyramids with flipping for both datasets. Specially, we follow previous works~\cite{cao2017realtime,newell2016associative} to refine estimation results with a single-person model trained on the same dataset on MPII and MSCOCO. 

\subsection{Results on MPII dataset}

\begin{table}[t!]\footnotesize 
	\caption{Comparison with state-of-the-arts on the full testing set of MPII dataset (mAP).}\label{tab:exp_mpii_sota}
	\centering
	\setlength{\tabcolsep}{1pt}
	\begin{tabular}{lccccccc>{\columncolor[gray]{0.9}}c>{\columncolor[gray]{0.9}}c}
		\toprule
		Method   &Head & Sho. & Elb. & Wri. & Hip & Knee  & Ank. & Total & Time[s]\\
		\midrule
		Iqbal and Gall~\cite{iqbal2016multi} & 58.4  & 53.9  & 44.5  & 35.0  & 42.2  & 36.7 & 31.1 & 43.1 & 10\\
		Insafutdinov \emph{et al.}~\cite{hpe:deepercut_eccv16} & 78.4  & 72.5  & 60.2  & 51.0  & 57.2  & 52.0 & 45.4 & 59.5 & 485\\
		Levinkov \emph{et al.}~\cite{levinkov2017joint} & 89.8 & 85.2 &	71.8 &	59.6 &	71.1 &	63.0 &	53.5 &	70.6 & - \\
		Insafutdinov \emph{et al.}~\cite{insafutdinov2016articulated} & 88.8  & 87.0  & 75.9  & 64.9  & 74.2  & 68.8 & 60.5 & 74.3 & - \\
		Cao \emph{et al.}~\cite{cao2017realtime} & 91.2  & 87.6  & 77.7  & 66.8  & 75.4  & 68.9 & 61.7 & 75.6 & 0.6\\
		Fang \emph{et al.}~\cite{fang16rmpe}& 88.4  & 86.5  & 78.6  & 70.4  & 74.4  & 73.0 & 65.8 & 76.7 & 0.4\\
		Newell and Deng~\cite{newell2016associative} & \textbf{92.1} & 89.3 & 78.9 & 69.8 & 76.2 & 71.6 & 64.7 & 77.5 & 0.25 \\
		Fieraru \emph{et al.}~\cite{fieraru2018learning} & 91.8	& \textbf{89.5} & 80.4 &	69.6 &	\textbf{77.3} &	71.7 &	65.5 &	78.0 & - \\
		\midrule
		SPM (Ours) & 89.7  & 87.4  & \textbf{80.4}  & \textbf{72.4}  & 76.7  & \textbf{74.9} & \textbf{68.3} & \textbf{78.5} & \textbf{0.058}\\
		\bottomrule
	\end{tabular}
\end{table}

\paragraph{Comparison with state-of-the-arts} In Table~\ref{tab:exp_mpii_sota}, we compare our SPM model with hierarchical SPR to state-of-the-arts on the full test split of MPII dataset\footnote{For our SPM model, the time is counted with single-scale testing on GPU TITAN X and CPU Intel I7-5820K 3.3GHz, excluding the refinement time by single-person pose estimation. For time evaluation on~\cite{newell2016associative}, we report the runtime with the code provided by authors in the link: https://github.com/umich-vl/pose-ae-train. For runtime on~\cite{cao2017realtime}, we refer to its speed for single-scale inference setting on MPII testing set, which can be found in Table 1 of 1st version of~\cite{cao2017realtime}.}. We can see that our SPM model only requires 0.058s to process an image, about $5{\times}$ faster than the bottom-up model~\cite{newell2016associative} with state-of-the-art speed, verifying the efficiency advantage of the proposed single-stage solution over existing two-stage ones for multi-person pose estimation. In addition, our SPM model achieves new state-of-the-art $78.5\%$ mAP on MPII dataset and improves accuracies for most kinds of body joints, which demonstrates its superior performance for estimating human poses of multiple persons in a single stage.

\begin{table}[t!]\footnotesize 
	\caption{Ablation experiments on MPII validation dataset (mAP).}\label{tab:exp_mpii_ablation}
	\centering
	\setlength{\tabcolsep}{2pt}
	\begin{tabular}{lccccccc>{\columncolor[gray]{0.9}}c>{\columncolor[gray]{0.9}}c}
		\toprule
		Method   &Head & Sho. & Elb. & Wri. & Hip & Knee  & Ank. & Total & Time[s]\\
		\midrule
		SPM-Vanilla & 91.7 & 87.5 & 76.1 & 65.2 & 75.2 & 71.4 & 60.3 & 75.3 & 0.058 \\
		SPM-Hierar & 92.0 & 88.5 & 78.6 & 69.4 & 77.7 & 73.8 & 63.9 & 77.7 & 0.058 \\ 
		\bottomrule
	\end{tabular}
\end{table}

\begin{figure}[t!]
	\begin{center}
		\includegraphics[scale=0.6]{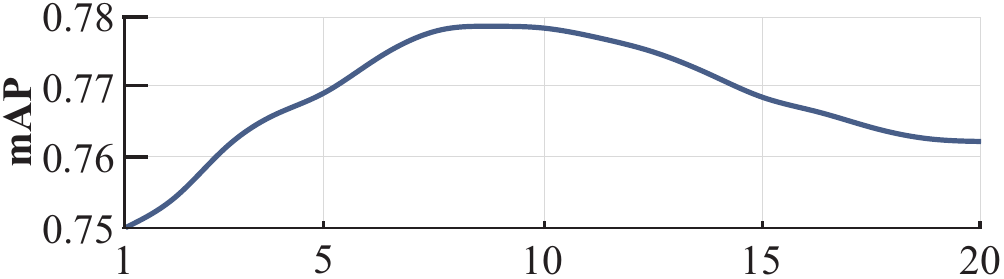}
		\caption{Analysis on hyper-parameter $\tau$, the neighborhood size for constructing regression target for body joint displacement.}
		\label{fig:exp_on_tau_and_nz}
	\end{center}
	\vspace{-3mm}
\end{figure}

\paragraph{Ablation analysis} We conduct ablation analysis on MPII validation dataset. We first evaluate the impact of the hierarchical division to SPR on the proposed SPM model.
Results are shown in Table~\ref{tab:exp_mpii_ablation}.
We use SPM-Vanilla and SPM-Hierar to denote the models for predicting SPR and Hierarchical SPR, respectively. 

We can see  SPM-Vanilla achieves $75.3\%$ mAP with 0.058s per image. By introducing joint hierarchies, SPM-Hierar improves the performance to $77.7\%$ mAP without increasing time cost as  SPR and  hierarchical SPR have the same complexity  and both of them are generated by SPM in a single-stage manner. In addition, we can see SPR-Hierar improves the accuracy of all joints. 
Moreover, we can also see that improvements by SPM-Hierar on long-range body joints wrists and ankles are significant, from $65.2\%$ to $69.4\%$ mAP and $60.3\%$ to $63.9\%$ mAP, respectively, verifying the effectiveness of shortening long-range displacements with Hierarchical SPR that divides body joints to different hierarchies.  These results clearly show the efficacy of incorporating hierarchical SPR to improve performance and efficiency of multi-person pose estimation.

We then conduct experiments to analyze the impact of important hyper-parameter $\tau$, the neighborhood size in constructing regression targets for body joint displacements in Section~\ref{sec:reg_target}, on the proposed SPM model.
We range $\tau$ from 1 to 20 and results are given in Figure~\ref{fig:exp_on_tau_and_nz}. From Figure~\ref{fig:exp_on_tau_and_nz}, we can see increasing $\tau$ from 1 to 7 gradually improves the performance, mainly because with the increase of positive samples, more variations of body joints can be covered for displacement regression in training. Further increasing $\tau$ from 7 to 10 cannot achieve performance improvement. However, when $\tau{>}10$, we observe performance drop. This is because noise from background is taken as positive samples and the overlap of displacement fields among multiple persons degrades the performance. Hence, we set $\tau{=}7$ in our experiments for the trade-off of efficiency and accuracy.

\begin{figure*}[h!]
	\begin{center}
		\includegraphics[scale=0.59]{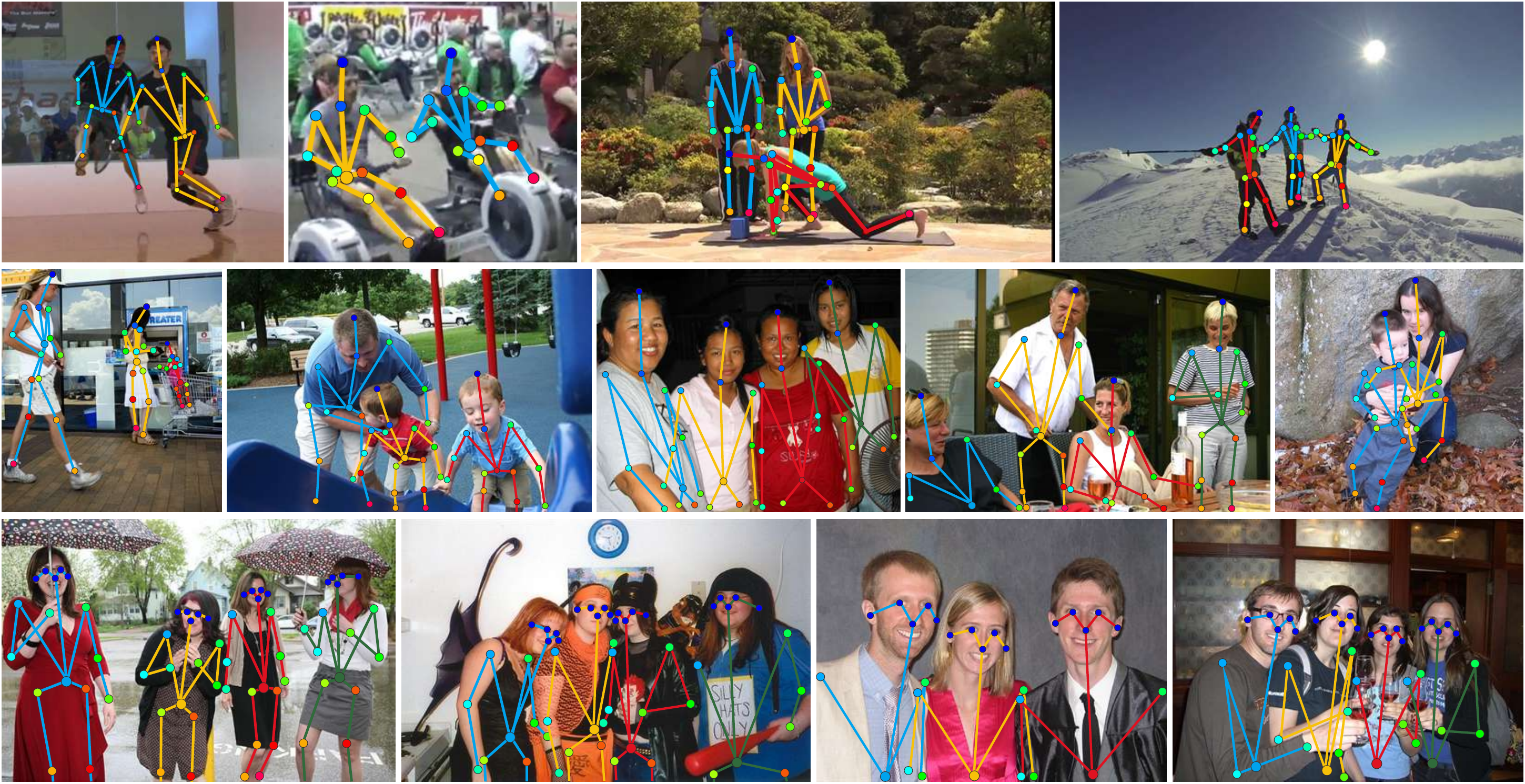}
		\caption{Qualitative results on MPII dataset (top), extended PASCAL-Person-Part dataset (middle) and MSCOCO dataset (bottom).}
		\label{fig:vis_examples}
	\end{center}
\end{figure*}

\begin{figure*}[h!]
	\begin{center}
		\includegraphics[scale=0.8]{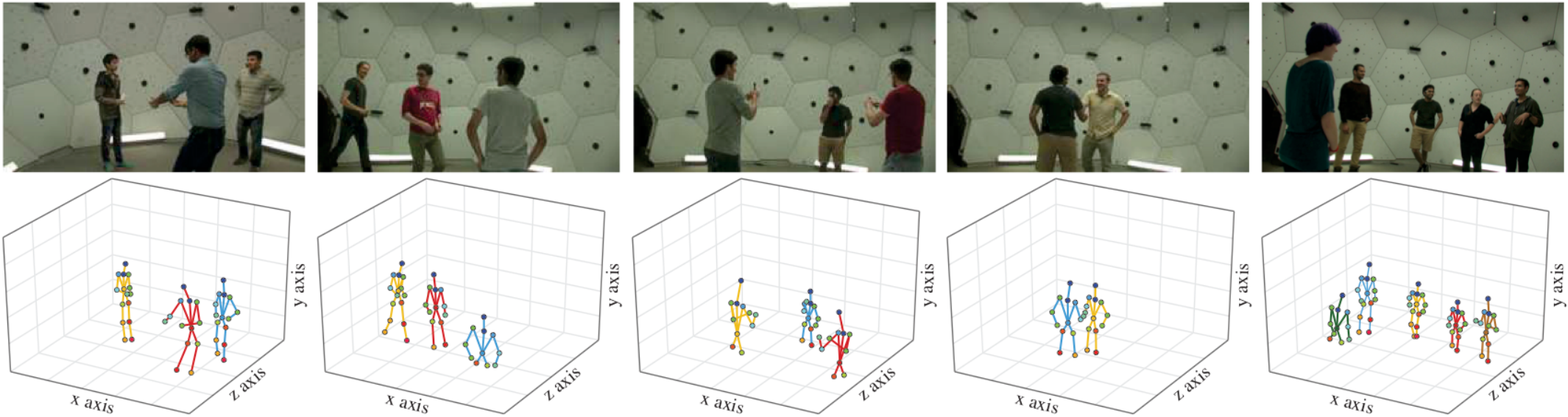}
		\caption{Qualitative results on CMU Panoptic dataset. 1st row is the input image and 2nd row is the corresponding multi-person 3D pose estimation with the proposed SPM. Best viewed in color and $2{\times}$ zoom.}
		\label{fig:3d_vis_examples}
	\end{center}
	\vspace{-3mm}
\end{figure*}

\vspace{-3mm}
\paragraph{Qualitative results} Qualitative results on MPII dataset are shown in the top row of Figure~\ref{fig:vis_examples}. We can see that the proposed SPM is effective and robust for estimating human poses represented by Hierarchical SPRs even in challenging scenarios, \eg, large pose deformation (1st example), blurred and cluttered background (2nd example), occlusion and person overlapping (3rd example), and illumination variations (4th example). These results further validate the efficacy of SPM.

\begin{table}[t!]\footnotesize
	\centering
	\caption{Comparison with state-of-the-arts on the testing set of the extended PASCAL-Person-Part dataset (mAP)}\label{tab:exp_pascal_sota}
	\centering
	\setlength{\tabcolsep}{2pt}
	\begin{tabular}{lccccccc>{\columncolor[gray]{0.9}}c}
		\toprule
		Method &Head & Sho. & Elb. & Wri. & Hip & Knee  & Ank. & Total\\
		\midrule
		Chen and Yuille~\cite{chen2015parsing} & 45.3 & 34.6 & 24.8 & 21.7 & 9.8 & 8.6 & 7.7 &  21.8\\
		Insafutdinov et al.~\cite{hpe:deepercut_eccv16} & 41.5 & 39.3 & 34.0 & 27.5 & 16.3 & 21.3 & 20.6 &  28.6\\
		Xia et at.~\cite{xia2017joint} & 58.0 & 52.1 & 43.1 & 37.2 & 22.1 & 30.8 & 31.1 & 39.2\\
		\midrule
		SPM (Ours) & \textbf{65.4} & \textbf{60.8} & \textbf{50.2} & \textbf{47.7} & \textbf{29.0} & \textbf{35.3} & \textbf{34.6}  & \textbf{46.1} \\
		\bottomrule
	\end{tabular}
	\vspace{-2mm}
\end{table}

\subsection{Results on PASCAL-Person-Part dataset}
Table~\ref{tab:exp_pascal_sota}  shows the comparison results with state-of-the-arts on the extended PASCAL-Person-Part dataset. We can see that the proposed SPM model achieves $46.1\%$ mAP and provides new state-of-the-art. Besides, SPM outperforms previous models for all body joints, demonstrating the effectiveness of the proposed single-stage model for tackling the multi-person pose estimation problem. 

Qualitative results are shown in the middle row of Figure~\ref{fig:vis_examples}. We observe SPM can deal with person scale variations (1st example), occlusion (2nd to 4th examples) and person overlapping (the last example), showing  the efficacy of SPM on producing robust pose estimation in various challenging scenes. 

\begin{table}[t!]\footnotesize
	\centering
	\caption{Comparison with state-of-the-arts on the MSCOCO test-dev (AP).}\label{tab:exp_coc_sota}
	\setlength{\tabcolsep}{2.5pt}
	\begin{tabular}{l>{\columncolor[gray]{0.9}}ccccc>{\columncolor[gray]{0.9}}c}
		\toprule
		Method   & AP & $\mathrm{AP}^{50}$ & $\mathrm{AP}^{75}$ & $\mathrm{AP}^{M}$ & $\mathrm{AP}^{L}$ & Time[s]  \\
		\midrule
		CMU-Pose~\cite{cao2017realtime} & 0.618 & 0.849 & 0.675 & 0.571 & 0.682 & 0.6\\
		RMPE~\cite{fang16rmpe} & 0.618 & 0.837 & 0.698 & 0.586 & 0.676 & 0.4\\
		Mask-RCNN~\cite{he2017mask} & 0.627 & 0.870 & 0.684 & 0.574 & 0.711 & 0.2\\
		G-RMI~\cite{papandreou2017towards} & 0.649 & 0.855 & 0.713 & 0.623 & 0.700 & - \\
		AssocEmbedding~\cite{newell2016associative} & 0.655 & 0.868 & 0.723 & 0.606 & 0.726 & 0.25\\
		PersonLab~\cite{papandreou2018personlab} & 0.687 & 0.890 & 0.754 & 0.641 & 0.755 & 0.464 \\
		\midrule
		SPM (Ours) & 0.669 & 0.885 & 0.729 & 0.626 & 0.731 & \textbf{0.058}\\
		\bottomrule
	\end{tabular}
\end{table}

\subsection{Results on MSCOCO dataset} 

Table~\ref{tab:exp_coc_sota} shows experimental results on MSCOCO test-dev. 
We can see that the proposed SPM model achieves overall 0.669 AP, which is slightly lower than the state-of-the-art~\cite{papandreou2018personlab}. However, our SPM achieves superior speed, $8\times $ faster than~\cite{papandreou2018personlab}. 
These results  further confirm the superior efficiency of our single-stage solution over existing two-stage top-down or bottom-up strategies, while achieving very competitive performance, 
for addressing the multi-person pose estimation tasks. 

Qualitative results on MSCOCO dataset are shown in the bottom row of Figure~\ref{fig:vis_examples}. We can see that our SPM model is effective in challenging scenes, \eg, appearance variations (1st example) and occlusion (2nd to 4th examples).

\subsection{Results on CMU Panoptic dataset}

We evaluate the proposed SPM model for multi-person 3D pose estimation on the CMU Panoptic dataset, which provides  large-scale data with accurate 3D pose annotations and thus is suitable to be an evaluation benchmark. Since previous works~\cite{Joo_2017_TPAMI,dong2019fast} only conduct  qualitative evaluation on this dataset, there are no reported  quantitative results  for  comparison. For better understanding the model performance, we present the first  quantitative evaluation here.  We separate 10,000 images  from  the dataset to form the testing split and use the remaining for training
as mentioned in Section~\ref{sec:exp_setup}. In particular, our SPM model achieves $77.8\%$ 3D-PCK, a promising result for multi-person 3D pose estimation. The effectiveness of our SPM model can be also verified through the qualitative results in Figure~\ref{fig:3d_vis_examples}. We can see our SPM model is robust for pose variations (1st and 2nd examples), self occlusions (3rd example), scale and depth changes (4th and 5th examples).

In addition, the proposed SPM model achieves attractive efficiency with speed of about 20 FPS. Moreover, its single-stage design also significantly simplifies the pipeline for multi-person 3D pose estimation from a single monocular RGB image, alleviating the requirements of intermediate 2D pose estimations~\cite{mehta2018single} or 3D pose reconstructions from multiple views~\cite{dong2019fast}. 

\section{Conclusion}

In this paper, we present the first single-stage model, Single-stage multi-person Pose Machine (SPM), for multi-person pose estimation. The SPM model offers a more compact pipeline and attractive efficiency advantage over existing two-stage based solutions. The superiority of SPM mainly comes from a novel Structured Pose Representation (SPR) that unifies the person instance and body joint position information and overcomes the intrinsic limitations of conventional pose representations. In addition, we present a hierarchical extension of SPR to effectively factorize long-range displacements into accumulative  short-range ones  between adjacent articulated joints, without introducing extra complexity to SPR. With SPR, SPM can estimate poses of multiple persons in a single-stage feed-forward manner. We implement SPM with CNNs, which can perform end-to-end learning and inference. Moreover, SPM can be flexibly adopted in both 2D and 3D scenarios. Extensive experiments on 2D benchmarks demonstrate the state-of-the-art speed of the proposed SPM model also with superior performance for predicting  poses of multiple persons. Results on 3D benchmark also show the promising performance of our SPM model with attractive efficiency.

\section*{Acknowledgement}
Jiashi Feng was partially supported by NUS IDS R-263-000-C67-646,  ECRA R-263-000-C87-133 and MOE Tier-II R-263-000-D17-112.

{\small
	\bibliographystyle{ieee_fullname}
	\bibliography{spm}
}

\end{document}